\documentclass{article} 
\usepackage{iclr2026_conference,times}

\usepackage{amsmath,amsfonts,bm}

\def\eqref#1{equation~\ref{#1}}

\def\1{\bm{1}}

\DeclareMathAlphabet{\mathsfit}{\encodingdefault}{\sfdefault}{m}{sl}
\SetMathAlphabet{\mathsfit}{bold}{\encodingdefault}{\sfdefault}{bx}{n}

\usepackage{hyperref}
\usepackage{url}
\usepackage{hyperref}
\usepackage{url}
\usepackage{booktabs}
\usepackage{multirow}
\usepackage{makecell}
\usepackage[dvipsnames]{xcolor}
\usepackage{graphicx}  
\usepackage{array} 
\usepackage{siunitx}
\usepackage{enumitem}
\usepackage{amsmath}
\usepackage{amssymb}
\usepackage{caption}
\usepackage{subcaption}

\usepackage[textsize=scriptsize]{todonotes}
\title{Knowledge-Augmented Long-CoT Generation \\ for Complex Biomolecular Reasoning}

\author{Tianwen Lyu\textsuperscript{1,3} \quad 
Xiang Zhuang\textsuperscript{2,3} \quad 
Keyan Ding\textsuperscript{3} \quad 
Xinzhe Cao\textsuperscript{4} \\
\textbf{Lei Liang\textsuperscript{5} \quad 
Wei Zhao\textsuperscript{6} \quad 
Qiang Zhang\textsuperscript{7}\thanks{Corresponding author.} \quad 
Huajun Chen\textsuperscript{2,3}\footnotemark[1]} \\
\textsuperscript{1}The Polytechnic Institute, Zhejiang University \\
\textsuperscript{2}College of Computer Science and Technology, Zhejiang University \\
\textsuperscript{3}ZJU-Hangzhou Global Scientific and Technological Innovation Center, Zhejiang University \\
\textsuperscript{4}University of Oxford \quad 
\textsuperscript{5}Ant Group \quad 
\textsuperscript{6}University of Aberdeen \\
\textsuperscript{7}ZJU-UIUC Institute, Zhejiang University \\
\texttt{\{22460438, zhuangxiang, qiang.zhang.cs, huajunsir\}@zju.edu.cn}
}

\iclrfinalcopy 
\begin{document}

\maketitle

\begin{abstract}
Understanding complex biomolecular mechanisms requires multi-step reasoning across molecular interactions, signaling cascades, and metabolic pathways. While large language models (LLMs) show promise in such tasks, their application to biomolecular problems is hindered by logical inconsistencies and the lack of grounding in domain knowledge. Existing approaches often exacerbate these issues: reasoning steps may deviate from biological facts or fail to capture long mechanistic dependencies. To address these challenges, we propose a Knowledge-Augmented Long-CoT Reasoning framework that integrates LLMs with knowledge graph–based multi-hop reasoning chains. The framework constructs mechanistic chains via guided multi-hop traversal and pruning on the knowledge graph; these chains are then incorporated into supervised fine-tuning to improve factual grounding and further refined with reinforcement learning to enhance reasoning reliability and consistency. Furthermore, to overcome the shortcomings of existing benchmarks, which are often restricted in scale and scope and lack annotations for deep reasoning chains, we introduce PrimeKGQA, a comprehensive benchmark for biomolecular question answering.
Experimental results on both PrimeKGQA and existing datasets demonstrate that although larger closed-source models still perform well on relatively simple tasks, our method demonstrates clear advantages as reasoning depth increases, achieving state-of-the-art performance on multi-hop tasks that demand traversal of structured biological knowledge. These findings highlight the effectiveness of combining structured knowledge with advanced reasoning strategies for reliable and interpretable biomolecular reasoning.
\end{abstract}

\section{Introduction}

Biological systems are governed by extraordinarily complex mechanisms spanning multiple levels, from protein–protein interactions to intracellular signaling cascades and metabolic pathways~\citep{stoney2018mapping}. Understanding these mechanisms is crucial for areas such as drug discovery, elucidation of disease mechanisms, and analysis of molecular interactions. Crucially, molecular events rarely occur in isolation; instead, they operate within intricate causal chains, making multi-step reasoning indispensable for interpretable mechanistic insights and scientific discovery~\citep{patel2005thinking, xu2024towards}.  
Meanwhile, the rapid expansion of genomic and proteomic datasets~\citep{ma2023retrieved, hosseini2024text2protein,fallahpour2025bioreason}, poses unprecedented challenges for computational methods to extract reliable and interpretable reasoning from large-scale biological data.

Recent advances in Large Language Models (LLMs)~\citep{hurst2024gpt, guo2025deepseek, yang2025qwen3} have recently achieved remarkable progress in multi-step reasoning tasks, particularly in domains such as mathematics~\citep{yu2023metamath}, logic~\citep{chen2024huatuogpt}, and programming~\citep{el2025competitive}. These advances are largely attributed to Chain-of-Thought (CoT)~\citep{wei2022chain, wang2022self, kojima2022large, min2024imitate} prompting techniques and reinforcement learning strategies, which enable stepwise and logically coherent reasoning. These advances highlight the potential of LLMs to support mechanistic reasoning in biomolecular research, where causal dependencies and multi-hop knowledge traversal are central. 

\begin{figure}[t]
    \centering    \includegraphics[width=0.9\linewidth, trim=120 260 470 110, clip]{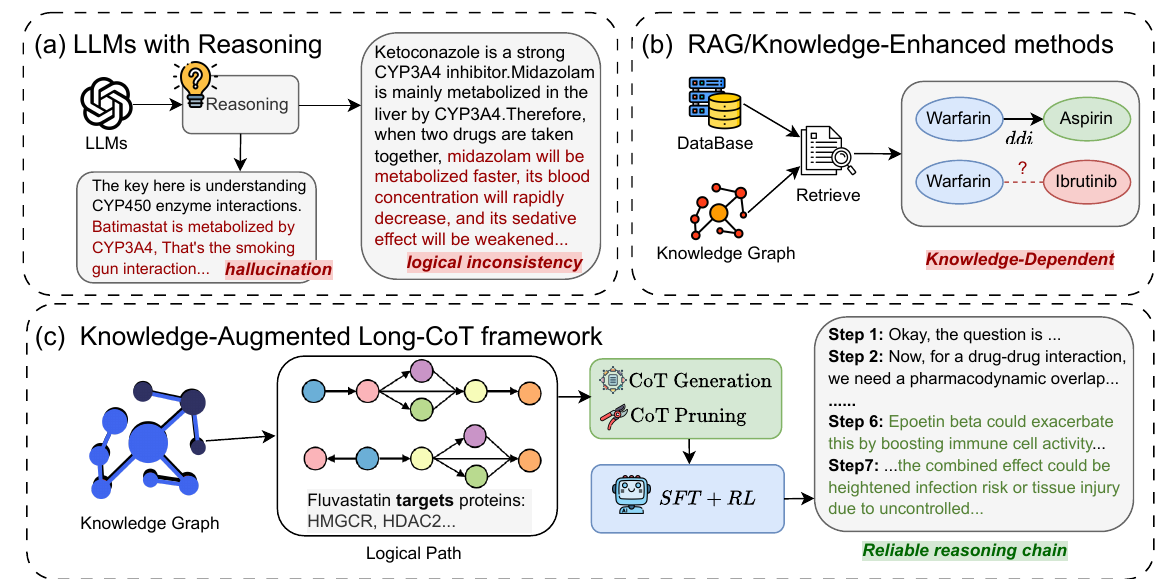}
    \caption{Comparison of different reasoning approaches for biomolecular problems. 
    (a) Reasoning LLMs generate multi-step reasoning chains but often suffer from hallucinations and logical inconsistencies. 
    (b) Retrieval-augmented generation and related knowledge-enhanced methods reduce hallucinations but are knowledge-dependent on the quality and coverage of sources. 
    (c) Our proposed Knowledge-Augmented Long-CoT framework integrates knowledge graph-guided reasoning, enabling logically coherent and reliable reasoning chains for complex biomolecular tasks.}
\label{fig:framework_comparison}
\end{figure}

However, direct application of LLMs to biomolecular reasoning remains highly non-trivial~\citep{li2024progress,zhuang2025advancing}: models can produce biologically implausible outputs~\citep{zheng2024large, fang2023domain}, generate reasoning chains with logical inconsistencies~\citep{zhang2024scientific}, and fail to leverage structured knowledge inherent in biological systems (Figure~\ref{fig:framework_comparison}a). Although recent knowledge-guided generation methods, such as retrieval-augmented~\citep{soman2024biomedical, li2025biomedrag} or knowledge-graph-based approaches~\citep{liu2021neural, luo2024graph,zhao2025biomaze}, partially alleviate hallucination and factual errors, but are inherently knowledge-dependent, relying on the coverage and reliability of external sources (Figure~\ref{fig:framework_comparison}b). \textbf{Addressing these limitations requires a framework that prioritizes the principled utilization of existing structured knowledge to guide model reasoning, rather than the pursuit of knowledge completeness, thereby enhancing LLM inference and enabling reliable, domain-grounded conclusions.}

To address these challenges, we propose Bio-KCoT, a knowledge-augmented long-CoT reasoning framework for complex biomolecular problems (Figure~\ref{fig:framework_comparison}c). At its core, Bio-KCoT introduces a knowledge graph–guided reasoning method that integrates structured biological knowledge to generate biologically plausible and logically coherent reasoning paths. Because raw knowledge graphs contain many redundant or spurious connections, we employ systematic path search and pruning to retain only reasoning chains that are both biologically meaningful and logically complete. These curated chains are then incorporated into supervised fine-tuning to strengthen factual grounding 
and reinforcement learning with Group Relative Policy Optimization (GRPO)
to refine reasoning strategies and enhance robustness on challenging multi-hop tasks. Finally, to enable systematic evaluation, we introduce PrimeKGQA, a benchmark derived from the PrimeKG knowledge graph~\citep{chandak2023building}, which spans diverse biomolecular question answering scenarios and task types. PrimeKGQA provides a standardized basis for fair comparison against strong baselines and supports progress in biologically grounded multi-step reasoning.
We summarize our contributions as follows:
\begin{itemize}
    \item 
    We introduce Bio-KCoT, a knowledge-augmented long-CoT reasoning framework for biomolecular problems. It integrates KG-guided path search and pruning 
    to generate coherent reasoning chains used during supervised fine-tuning and reinforcement learning.
    \item We introduce a high-quality benchmark for biomolecular reasoning, built by systematically collecting and curating knowledge graph-guided reasoning paths that capture multi-level semantic relationships. To ensure fair evaluation, the dataset encompasses diverse biomolecular reasoning scenarios and task types.
    \item Experimental results on both PrimeKGQA and existing datasets show that while larger closed-source models may retain advantages on simpler tasks, our method achieves substantial gains as reasoning complexity increases, delivering the strongest performance on multi-hop tasks requiring traversal of structured biological knowledge.
\end{itemize}

\section{Related work}
\label{others}
In this section, we provide a comprehensive review of related work, highlighting prior approaches and their relevance to biomolecular reasoning.

\subsection{Chain-of-Thought reasoning}
Chain-of-Thought reasoning enables LLMs to generate intermediate reasoning steps before arriving at a final answer~\citep{wei2022chain, kojima2022large}. This paradigm not only improves the transparency of model predictions, but also enhances performance on tasks requiring multi-step reasoning. Early work introduced few-shot CoT prompting with manually crafted demonstrations~\citep{wei2022chain}, or simple zero-shot prompting such as ``Let’s think step by step''~\citep{kojima2022large}. To further strengthen the quality of reasoning, several extensions have been proposed, including self-consistency~\citep{wang2022self}, least-to-most prompting~\citep{zhou2022least}, complexity-based prompting~\citep{fu2022complexity}, and self-polishing techniques~\citep{xi2023self}.
More recently, CoT reasoning has been explicitly incorporated into the design of advanced reasoning-oriented LLMs, leading to the emergence of models such as o1~\citep{openaio1}, DeepSeek-R1~\citep{guo2025deepseek}, and Google’s Gemini 2.5 Pro~\citep{comanici2025gemini}, which integrate dedicated “thinking” mechanisms and training strategies to enhance multi-step reasoning, coding, and scientific problem solving.
Concurrently, the emergence of reasoning-oriented LLMs stimulated interest in distilling their capabilities into smaller student models ($<$10B), where teacher-generated CoTs are employed as effective supervision signals~\citep{ho2022large, fu2023specializing, magister2022teaching}.
However, distilled models still struggle to generalize reasoning beyond benchmark datasets, particularly when factual grounding is critical.

\subsection{Scientific and Biomolecular CoT Reasoning}
Beyond general NLP benchmarks, the CoT paradigm is increasingly being adapted for complex reasoning in scientific domains. In medicine, models like HuatuoGPT~\citep{chen2024huatuogpt} and ReasonMed~\citep{sun2025reasonmed} leverage multi-stage, knowledge-infused pipelines to enhance diagnostic reliability~\citep{muennighoff2025s1, dutta2024adaptive, zuo2025kg4diagnosis}. This trend is mirrored in the biomolecular sciences, where models are being developed to reason about intricate biological and chemical processes. These efforts range from incorporating fundamental molecular features like DNA sequences~\citep{fallahpour2025bioreason}, protein evolutionary profiles~\citep{liu2025proteinreasoner}, and molecular structures~\citep{m2024augmenting, jang2024chain, narayanan2025training}, to modeling higher-level systems such as protein-protein interaction pathways~\citep{jin2024prollm}, tree-structured biological processes~\citep{hsu2024thought}, and single-cell annotations~\citep{fang2025cell}.

To facilitate progress, the community relies on benchmarks such as ScienceQA~\citep{lu2022learn}, MedQA~\citep{jin2021disease}, PubMedQA~\citep{jin2019pubmedqa}, and BioASQ~\citep{krithara2023BioASQ}. More recently, pathway-centric datasets have been introduced to specifically challenge reasoning over biochemical interactions~\citep{li2023chatpathway, park2025comparative, zhao2025biomaze}. However, these existing resources often lack the explicit, multi-step rationales tailored to biomolecular reasoning. This scarcity of detailed ground-truth reasoning chains limits their effectiveness in training the very long-CoT models needed to tackle deep biological questions.

\subsection{Knowledge-Augmented Reasoning}
Although CoT provides a reasoning scaffold, its reliability is fundamentally limited by the internal knowledge of the LLM, which can be prone to hallucination~\citep{huang2025survey,griot2025large, chen2024cod}. 
To address this limitation, recent work has explored knowledge augmentation, which grounds reasoning in external verified sources such as domain-specific databases or KGs~\citep{chen2024huatuogpt, xie2024medtrinity}. Among these, KGs are particularly powerful as they provide structured knowledge about entities and their relationships, which can be integrated into LLM reasoning through various techniques, from embedding-based methods to modern hybrid frameworks that enable direct interaction between the LLM and the graph~\citep{guo2024mkgl, wang2023biobridge, chandak2023building}. {This ability to represent complex relational structures makes KGs particularly valuable in domains such as biomedicine, which are defined by intricate networks of interactions.}

For biomolecular reasoning, where accurate modeling of interactions among genes, proteins, and pathways is essential, knowledge-augmented approaches are particularly promising~\citep{liu2021neural,luo2024graph,zhao2025biomaze}. While approaches such as MedReason~\citep{wu2025medreason} and KG-o1~\citep{wang2025kg} leverage KGs {for supervised fine-tuning through shortest-path–based extraction, this strategy is generally well-suited for tasks where relevant evidence can indeed be captured within short paths. However, its effectiveness diminishes for complex problems that require integrating knowledge from disparate parts of the graph. By constraining the evidence to a local neighborhood, it not only risks oversimplifying the reasoning process but also fails to support robust, multi-step inference.}

\begin{figure}[t]
    \centering    \includegraphics[width=\linewidth,trim=30 120 30 90,clip]{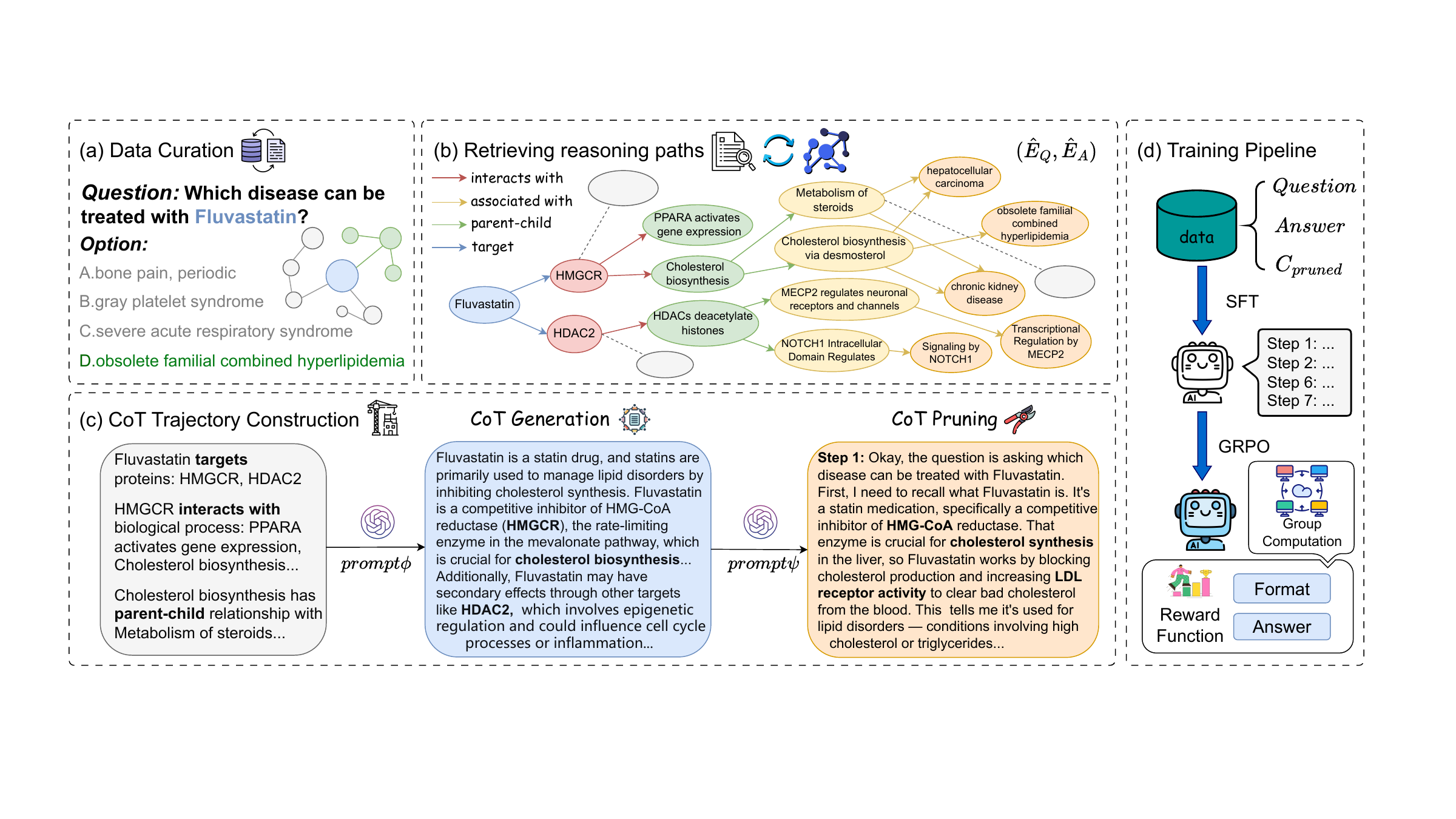}
    \caption{
    Overview of our proposed framework. 
    (a) \textbf{Data Curation}: Given a biomolecular question and candidate answers, we extract entities from the question ($E_Q$) and the correct answer ($E_A$). 
    (b) \textbf{Retrieving Reasoning Paths}: The extracted entities are mapped onto KG nodes. Reasoning paths $\mathcal{P}(Q,A;d)$ are then retrieved using predefined templates.
    (c) \textbf{CoT Trajectory Construction}:  The induced path $p$ provides semantic relations that guide the initial CoT generation. The generated trajectories are further refined through the pruning stage to ensure clarity and accuracy.
    (d) \textbf{Training Pipeline}: The curated $(Q,A,C_{\text{pruned}})$ pairs are used for supervised fine-tuning (SFT), followed by reinforcement learning (GRPO) to align the model’s reasoning and answer generation.
    }
    \label{fig:method}
\end{figure}

\section{Methodology}
\label{gen_inst}
The overall workflow, including data curation, reasoning path retrieval, CoT generation, and training pipeline, is illustrated in Figure~\ref{fig:method}.
\subsection{Retrieving reasoning paths}
\label{ssec:Retrieving}
Given a biomolecular question $Q$ (with problem statement and candidate options) and its corresponding correct answer $A$, our first step is to retrieve reasoning paths from the pre-constructed biomolecular KG $G=(V,E)$, where $V$ denotes the set of entities and $E$ the relations. The goal is to extract structured reasoning chains $\mathcal{P}$ that bridge the question and the answer.

\paragraph{Entity extraction and mapping.}
We extract entities mentioned in $Q$ and $A$, obtaining two sets:
\begin{equation}\label{eq:entity_sets}
E_Q = \{ e_i^Q \}_{i=1}^{n}, \quad 
E_A = \{ e_j^A \}_{j=1}^{m}.
\end{equation}
$E_Q$ and $E_A$ denote the sets of entities extracted from the question and the answer, respectively, where $n$ and $m$ are the corresponding numbers of entities.
These entities are then mapped to their corresponding nodes in the KG:
\begin{equation}\label{eq:mapping}
\hat{E}_Q = \{ \hat{e}_i^Q \mid e_i^Q \mapsto \hat{e}_i^Q \in V \}, 
\quad 
\hat{E}_A = \{ \hat{e}_j^A \mid e_j^A \mapsto \hat{e}_j^A \in V \}.
\end{equation}
$\hat{E}_Q$ and $\hat{E}_A$ represent the sets of nodes in the KG corresponding to the extracted entities.

\begin{table}[t]
\caption{Illustration of task categories at different difficulty levels.
Basic tasks rely on short, direct reasoning in the KG (e.g., drug indication, biological process). Medium tasks involve longer, contextualized reasoning across intermediate entities (e.g., Off-label use, disease–protein associations, side effects). Hard tasks require multi-hop integration of heterogeneous concepts (e.g., contraindications, drug–drug interactions).
Here, the blue node denotes the head entity in the question, and the orange node denotes the correct answer entity.
}
\label{tab:task-illustration-revised}
\centering
\newcommand{\includecompactgraphics}[1]{\raisebox{-.5\height}}
\begin{tabular}{l l m{4.5cm} >{\hspace{5mm}}m{3.4cm}}
\toprule
\textbf{Level} & \textbf{Task Category} & \textbf{Example} & \textbf{Illustration} \\
\midrule
\renewcommand{\arraystretch}{0.8}
\multirow{2}{*}{\textbf{Basic}} & Indication & Which \textbf{disease} can be treated with \textbf{Dalfampridine}? & \includegraphics[width=1.3\linewidth, trim=240 260 170 280, clip]{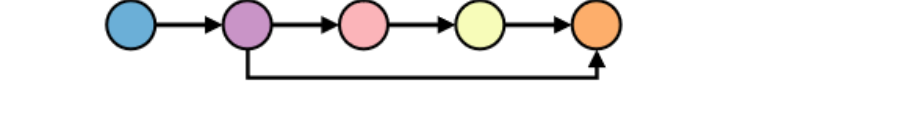} \\
\cmidrule(l){2-4}
& Bioprocess & Which \textbf{biological process} is associated with \textbf{LIMS1}? & 
\includegraphics[width=1.3\linewidth, trim=270 262 140 272, clip]{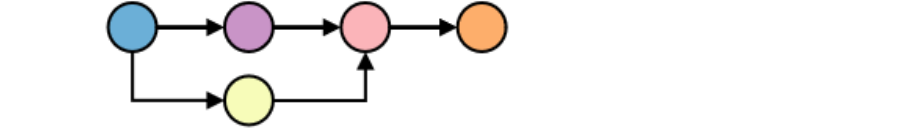} \\
\midrule 
\multirow{3}{*}[-3ex]{\textbf{Medium}} & Off-label Use & Which \textbf{drug} is used Off-label for \textbf{botulism}? & 
\raisebox{5pt}{\includegraphics[width=1.3\linewidth, trim=240 255 170 270, clip]{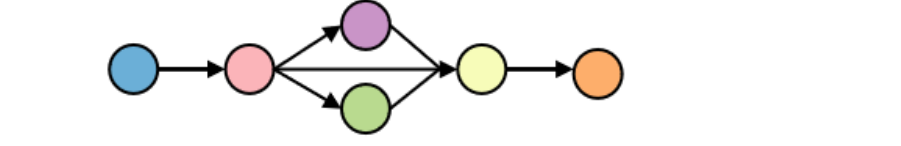}} \\
\cmidrule(l){2-4}
& Disease-Protein & Which \textbf{protein} is associated with \textbf{Cutrarino triad}? & \raisebox{5pt}{\includegraphics[width=1.3\linewidth,, trim=245 255 155 260, clip]{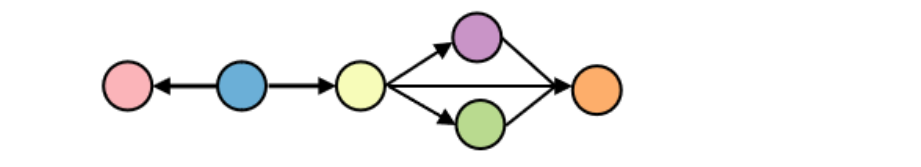}} \\
\cmidrule(l){2-4}
& Side effect & What is a known \textbf{side effect} of \textbf{Flurbiprofen}? & \raisebox{5pt}{\includegraphics[width=1.3\linewidth, trim=215 255 180 265, clip]{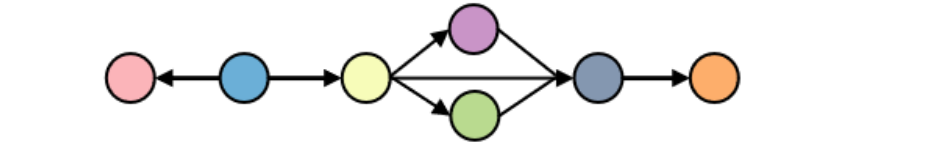}} \\
\midrule
\multirow{2}{*}{\textbf{Hard}} & Contraindication & Which \textbf{disease} is contraindication for \textbf{Nitrogen}? & \raisebox{5pt}{\includegraphics[width=1.3\linewidth, trim=220 255 160 255, clip]{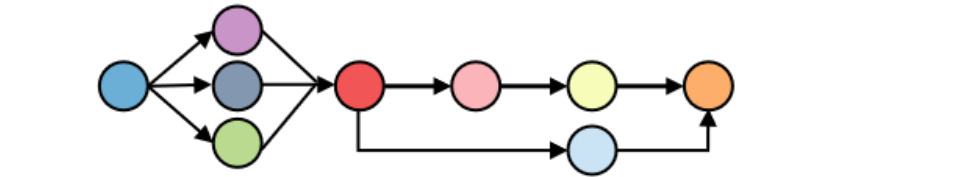}} \\
\cmidrule(l){2-4}
& Drug Drug Interaction & Which \textbf{drug} has a drug drug interaction with \textbf{Piritrexim}? & \raisebox{5pt}{\includegraphics[width=1.3\linewidth, trim=220 255 155 255, clip]{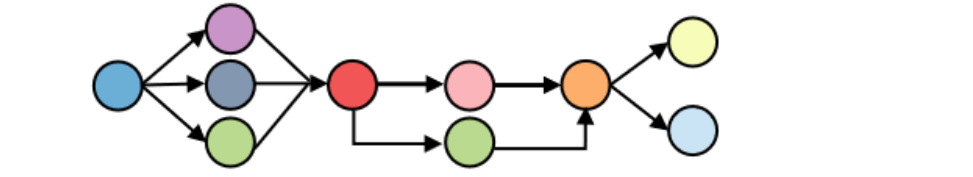}} \\
\bottomrule
\renewcommand{\arraystretch}{1.0}
\end{tabular}
\end{table}

\paragraph{Path definition and extraction.}
Reasoning paths are instantiated from a predefined set of abstract templates, denoted as $\mathbb{T} = \{\mathcal{T}_1, \mathcal{T}_2, \dots, \mathcal{T}_N\}$. Each template $\mathcal{T}_i \in \mathbb{T}$ encodes a specific topological structure of entity and relation transitions, designed to link a question entity $u \in \hat{E}_Q$ to a candidate answer entity $v \in \hat{E}_A$. 
These templates accommodate a spectrum of structural complexities, ranging from simple linear chains to intricate non-linear graphs. To illustrate the diversity of these templates, we define three fundamental structural patterns:

\begin{enumerate}[label=(\roman*), align=left]

    \item \emph{Linear chain structures}, where entities are connected by a single sequential relation chain:
    \begin{equation}
        \pi_{\mathrm{lin}}(u,v) = (u = v_0 \xrightarrow{r_1} v_1 \xrightarrow{r_2} v_2 \cdots 
        \xrightarrow{r_d} v_d = v).
    \end{equation}

    \item \emph{Divergent structures}, where multiple reasoning trajectories emanate from a common entity:
    \begin{equation}
        \pi_{\mathrm{div}}(u,v) = 
        \Big\{ (u=v_0 \xrightarrow{r_1} v_1),\; 
        (v_0 \xrightarrow{r_2} v_2 \xrightarrow{r_3} \cdots \xrightarrow{r_d} v_d=v) \Big\}.
    \end{equation}

    \item \emph{Convergent structures}, where distinct relational branches terminate at a common entity:
    \begin{equation}
        \pi_{\mathrm{con}}(u,v) = 
        \Big\{ (u=v_0 \xrightarrow{r_1} v_1 \xrightarrow{r_2} \cdots v_d=v),\;
        (v_0 \xrightarrow{r_3} v_2 \xrightarrow{r_4} \cdots v_d=v) \Big\}.
    \end{equation}

\end{enumerate}

In this formulation, $d$ denotes the overall reasoning complexity of a path, 
which jointly captures (i) the \emph{depth}, corresponding to the maximal length 
of a relational chain, and (ii) the \emph{breadth}, reflecting the number of admissible branches within the structure.

For a given template $\mathcal{T} \in \{\mathrm{lin},\mathrm{div},\mathrm{con}\}$, we define the set of all concrete reasoning paths derived from $\mathcal{T}$ as
\begin{equation}
\mathrm{Inst}(\mathcal{T}; u,v) \triangleq {\pi(u,v) \mid \pi(u,v) \models \mathcal{T} }.
\end{equation}

Accordingly, the set of all instantiated paths connecting the question $Q$ and 
candidate answers $A$ can be formally expressed as
\begin{equation}\label{eq:paths_set_combined}
\mathcal{P}(Q,A) = \bigcup_{u \in \hat{E}_Q,\, v \in \hat{E}_A} 
\Big\{ \pi(u,v) \;\big|\; 
\pi(u,v) \in \mathrm{Inst}(\mathcal{T}; u,v),\; \mathcal{T} \in \mathbb{T},\; v_0 = u,\; v_{|\pi|} = v \Big\},
\end{equation}
which collects all admissible paths derived from any template in $\mathbb{T}$ 
connecting entities in $Q$ to candidate answers in $A$.

\paragraph{Task difficulty characterized by hop depth.}
{We define task difficulty through the complexity metric $d$, which aggregates two key factors: the maximal depth of the relational chain and the breadth of its structural branches.}
Greater depths correspond to broader knowledge scopes and higher degrees of informational fragmentation, thereby requiring more demanding semantic integration and logical inference. Accordingly, task difficulty is stratified into discrete levels as a function of $d$ (Eq.~\ref{eq:difficulty}).{Table~\ref{tab:task-illustration-revised} further illustrates the mapping between representative task categories and their associated depth ranges.}
\vspace{-0.5em}
\begin{equation}\label{eq:difficulty}
\text{Task Difficulty} =
\begin{cases}
\text{Basic (e.g., Indication, Bioprocess)}, & d \le5,\\[2pt]
\text{Medium (e.g., Off-label use, Disease--Protein, Side effect)}, & 6 \le d \le 7,\\[2pt]
\text{Hard (e.g., Contraindication, Drug Drug Interaction)}, & d \ge 8.
\end{cases}
\end{equation}

In the next stage, these reasoning paths serve as the factual and structural foundation for guiding CoT generation.

\subsection{CoT Generation and Pruning}
\label{ssec:Generation}
Building upon the retrieved reasoning paths $P$, the next step is to generate a fact-grounded CoT that faithfully reflects the reasoning process underlying the question–answer pair. Each path $\pi \in \mathcal{P}$ provides a structured sequence of entity and relation transitions that can serve as a scaffold for the CoT, ensuring that the reasoning process is grounded in valid biomedical knowledge.

\paragraph{CoT generation.}
We define a prompt constructor $\Phi$ that incorporates the question $Q$, the correct answer $A$, the reasoning paths $\mathcal{P}$. The details of this prompt constructor is illustrated in Figure~\ref{fig:cot-generation} of Appendix~\ref{app:prompts}. Given this structured input, the LLM produces a reasoning chain, which we denote as:
\begin{equation}\label{eq:cot_gen}
C \;=\; \mathrm{LLM}\big( \Phi(Q,A,\mathcal{P}) \big).
\end{equation}
{Here, $C$ represents the initial CoT generated by the LLM. The generation process is guided by KG-derived evidence, ensuring that the generated CoT remains aligned with established biomedical knowledge and avoids unsupported associations.}

\paragraph{CoT pruning.}
To enhance the clarity of the factually-grounded CoT($C$), which may contain redundant steps or tangential details, 
we define a pruning prompt constructor $\Psi$ that takes as input the question $Q$ and the initial reasoning chain $C$. The details of $\Psi$ is illustrated in Figure~\ref{fig:cot-pruning}  of Appendix~\ref{app:prompts}. This prompt instructs the LLM to refine $C$ by removing unnecessary reasoning steps and preserving only the essential logical chain leading to the answer:
\begin{equation}\label{eq:pruned_via_llm}
C_{\mathrm{pruned}} \;=\; \mathrm{LLM}\big( \Psi(Q,C) \big).
\end{equation}
Here, $C_{\mathrm{pruned}}$ represents the refined reasoning chain that retains only the essential logical steps.
Through this prompt-based pruning strategy, the reasoning process becomes more concise and interpretable while maintaining accuracy. A concrete example is provided in Figure~\ref{fig:example} of Appendix~\ref{app:prompts}.

{In summary, by leveraging KG-derived paths $\mathcal{P}$, this two-stage process generates a concise and evidence-grounded reasoning chain $C_{\mathrm{pruned}}$, achieving both factual correctness and interpretability.}

\subsection{Supervised Fine-tuning and Reinforcement Learning}
After building the reasoning dataset, we conduct supervised finetuning and reinforcement learning.
Through the SFT stage, the base model (Qwen3-4B and Qwen3-8B) are trained to follow KG-guided templates, generate coherent multi-step reasoning, and ensure output validity for complex biomolecular questions. This stage effectively mitigates reward sparsity, establishing a robust foundation for the model to efficiently refine its reasoning capabilities during the reinforcement learning phase.

For the reinforcement learning phase, we adopt Group Relative Policy Optimization (GRPO)~\citep{shao2024deepseekmath}, which leverages intra-group advantage estimation to eliminate the need for a separate critic network. The training process for each input $x$ involves sampling $G$ candidate responses from the current policy and optimizing it using a clipped objective function with group-normalized advantages (the detailed formulation is provided in Appendix~\ref{app:grpo}).

To simultaneously promote structured reasoning and factual accuracy, we design a composite reward function that enforces both the output format and the correctness. Each response is required to follow a reasoning template, containing intermediate reasoning in \texttt{<think>...</think>} and the final answer in \texttt{<answer>...</answer>}. 
The reward is composed of two parts:
\begin{equation}
R_{\text{format}} = \begin{cases} 
1, & \text{if format is valid}, \\ 
0, & \text{otherwise}, 
\end{cases}
\quad
R_{\text{answer}} = \begin{cases} 
5, & \text{if the predicted answer is correct}, \\ 
0, & \text{otherwise}, 
\end{cases}
\end{equation}
and the total reward is
$R_{\text{reward}} = R_{\text{format}} + R_{\text{answer}}$.
This reward structure ensures that answer correctness grants a primary reward of $5$, while format adherence provides an additional reward of $1$. In this way, these incentives promote both syntactic validity and semantic accuracy, guiding the model toward reliable reasoning and outputs.

\section{Experiment and Results}
\label{headings}

\subsection{Experimental Setup}
\paragraph{Dataset and Benchmark}

\label{sec:dataset}
{Existing biomedical QA benchmarks (e.g., BioASQ\citep{krithara2023BioASQ}, BiomixQA\citep{soman2024biomedical}) remain limited in scope: they are relatively small and lack explicit annotations for multi-hop reasoning over structured knowledge. These shortcomings restrict their ability to evaluate the deep reasoning capacities of the models. In contrast,} we introduce PrimeKGQA, a benchmark constructed from PrimeKG~\citep{chandak2023building}. QA pairs are generated using a template-based approach: Each question is formed from a head–relation pair, with the corresponding tail entity serving as the answer (Figure~\ref{fig:method}a).
The tasks cover diverse biomedical categories(e.g., diseases, drugs, genes, pathways) and exhibit varying levels of reasoning difficulty. In particular, the path length in PrimeKG naturally reflects the complexity of the reasoning required to answer a question (Table~\ref{tab:task-illustration-revised}). To prevent data leakage, the dataset is partitioned by head entities, ensuring no overlap between training and test sets. In total, PrimeKGQA consists of 6,710 QA pairs, with 3,500 for supervised fine-tuning, 1,500 for reinforcement learning, and 1,710 for evaluation. {Detailed statistics of the test data are provided in the Appendix~\ref{app:datasets}. The QA pairs are in the form of single-choice questions with four options, and the evaluation metric is accuracy.}

\paragraph{Baselines}
For the empirical validation of our proposed data and training methodology, we conduct a comparative analysis against a comprehensive set of baselines. To enable a structured comparison, these baselines are grouped into three categories:
(1) Open-source model Qwen3-4B and Qwen3-8B~\citep{yang2025qwen3} 
{(base model baseline)}; (2) Closed-source models, including GPT-4o-mini, GPT-4o~\citep{hurst2024gpt}, and Gemini 2.5 Pro~\citep{comanici2025gemini}; (3) Reasoning-oriented models, such as o1~\citep{jaech2024openai}, o3-mini, and Deepseek-R1~\citep{guo2025deepseek}.

\paragraph{Training and Inference}
We perform further training on the Qwen3-4B and Qwen3-8B models~\citep{yang2025qwen3}. {The training process follows a two-stage paradigm, consisting of full finetuning and reinforcement learning(RL), using the PrimeKGQA dataset (see Section~\ref{sec:dataset}). In the finetuning stage, the model is trained for 4 epochs with a learning rate of $1.0 \times 10^{-5}$, a per-device batch size of 2, and a maximum input length of 4096 tokens. For the subsequent RL phase, we transition to a parameter-efficient approach, employing Low-Rank Adaptation (LoRA) with a rank of 32. The model undergoes one epoch of training with a learning rate of $1.0 \times 10^{-6}$, a per-device batch size of 4, and a maximum input length of 4096 tokens.} During inference, all models are prompted to generate CoT style responses.

\subsection{Main results}
\begin{table}[t]
\centering
\caption{Performance comparison of Open-source, Closed-source, Reasoning-oriented models, and our proposed model on the PrimeKGQA benchmark test set. The table shows accuracy scores across task categories of varying difficulty levels.
Higher values indicate better performance.}
\label{tab:model-performance-merged}
\setlength{\tabcolsep}{4pt} 
\resizebox{\textwidth}{!}{
\begin{tabular}{ll cc c ccc c cc c c}
\toprule
\multirow{2}{*}[-1.5ex]{\textbf{Type}} & \multirow{2}{*}[-1.5ex]{\textbf{Model}} & \multicolumn{3}{c}{\textbf{Basic}} & \multicolumn{4}{c}{\textbf{Medium}} & \multicolumn{3}{c}{\textbf{Hard}} & \multirow{2}{*}[-1.5ex]{\textbf{All Avg.}} \\
\cmidrule(lr){3-5} \cmidrule(lr){6-9} \cmidrule(lr){10-12}
& & \thead{Indi- \\ cation} & \thead{Bio- \\ process} & Avg. & \thead{Off-label \\ use} & \thead{Disease- \\ Protein} & \thead{Side \\ effect} & Avg. & \thead{Contra- \\ indication} & DDI & Avg. & \\
\midrule
\multirow{2}{*}{\makecell[l]{Open \\ Source}} & \makecell[l]{Qwen3-4B} & 0.768 & 0.523  & 0.606 & 0.710 & 0.573  & 0.616 & 0.618 & 0.438 & 0.503 & 0.473 & 0.568 \\
& \makecell[l]{Qwen3-8B} & 0.845 & 0.653 & 0.749 & 0.671 & 0.600 & 0.516 & 0.586 & 0.452 & 0.537 & 0.498 & 0.601 \\
\midrule
\multirow{3}{*}{\makecell[l]{Closed \\ Source}} & \makecell[l]{GPT-4o-mini} & 0.903 & 0.740 & 0.796 & \underline{0.903} & 0.713 & 0.844 & 0.801 & 0.564 & \underline{0.683} & 0.625 & 0.743 \\
& \makecell[l]{Gemini 2.5 Pro} & \underline{0.910} & 0.850 & {0.880} & 0.710 & 0.803 & 0.804 & 0.783 & 0.664 & 0.647 & 0.655 & 0.768 \\
& \makecell[l]{GPT-4o} & 0.890 & 0.873 & 0.879 & 0.897 & 0.807 & 0.844 & \underline{0.840} & 0.676 & 0.623 & 0.650 & 0.789 \\
\midrule
\multirow{3}{*}{\makecell[l]{Reasoning\\oriented}} 
& \makecell[l]{Deepseek-R1} & 0.897 & 0.777 & 0.837 & 0.839 & 0.750 & 0.803 & 0.788 & 0.532 & 0.625 & 0.583 & 0.735 \\
& \makecell[l]{o3-mini} & 0.897 & \underline{0.877} & \underline{0.884} & 0.852 & \textbf{0.857} & 0.772 & 0.826 & 0.580 & 0.663 & 0.625 & 0.777 \\
& \makecell[l]{o1} & \textbf{0.936} & \textbf{0.893} & \textbf{0.908} & 0.832 & \underline{0.840} & 0.844 & \underline{0.840} & 0.696 & \textbf{0.703} & \underline{0.700} & \underline{0.813} \\
\midrule
\multirow{2}{*}{\makecell[l]{Ours}} 
& \makecell[l]{\textbf{Bio-KCoT(4B)}} & 0.897  & 0.773  & 0.815 & 0.890 & 0.743    & \textbf{0.912} & 0.835 & \underline{0.804} & 0.610 & 0.698 & 0.786 \\ 
& \makecell[l]{\textbf{Bio-KCoT(8B)}} & 0.901  & 0.807 & 0.839 & \textbf{0.930} & 0.793    & \underline{0.908} & \textbf{0.864} & \textbf{0.848} & 0.643 & \textbf{0.736} & \textbf{0.816} \\

\bottomrule
\end{tabular}
}
\end{table}

Table~\ref{tab:model-performance-merged} reports the experimental results on the PrimeKGQA benchmark. On basic-level tasks, advanced closed-source LLMs still maintain a leading position, primarily due to their extensive parameter knowledge and strong factual memory capabilities. However, as task difficulty increases, this advantage gradually diminishes and is eventually surpassed. On medium-level tasks, the Bio-KCoT 8B model achieves an average score of 0.864, outperforming both o1 and GPT-4o, which each scored 0.840. Specifically, in the Off-label use subtask, Bio-KCoT 8B achieves 0.930, significantly higher than GPT-4o’s 0.897. In the side effect subtask, the 4B and 8B versions of Bio-KCoT achieve 0.912 and 0.908, respectively, far exceeding the 0.844 scores of GPT-4o and o1.

{On the most complex hard-level tasks, Bio-KCoT 8B continues to lead with an overall score of 0.736. Its advantage is more pronounced in the contraindication subtask, where it achieves 0.848, substantially ahead of o1 at 0.696. The experiments also show that Bio-KCoT’s performance improves with increased model size, demonstrating a favorable scale-up property. These results indicate that, with careful framework design and the incorporation of domain knowledge, even models with relatively small parameter counts can achieve performance on complex biomedical tasks that matches or exceeds that of some closed-source large models.}

{In addition, we observe that the performance of most baseline methods declines progressively from the basic to the medium and hard levels as task difficulty increases. In contrast, our method performs better on medium tasks than on basic ones, and its results on hard tasks remain comparable to those on the basic level. This suggests that the incorporation of long chains of thought becomes particularly advantageous as task complexity increases, enabling the model to leverage reasoning chains more effectively when shallow memorization is insufficient. By generating knowledge-guided reasoning chains, the model can gradually unfold intermediate steps and integrate structured knowledge with semantic information, thereby addressing high-complexity problems more effectively. To identify limitations for future improvement, see the error analysis in the Appendix~\ref{app:error}.}

\subsection{Generalization to other benchmarks}

\begin{figure}[t]
    \centering    \includegraphics[width=\linewidth, trim=0 620 0 10, clip]{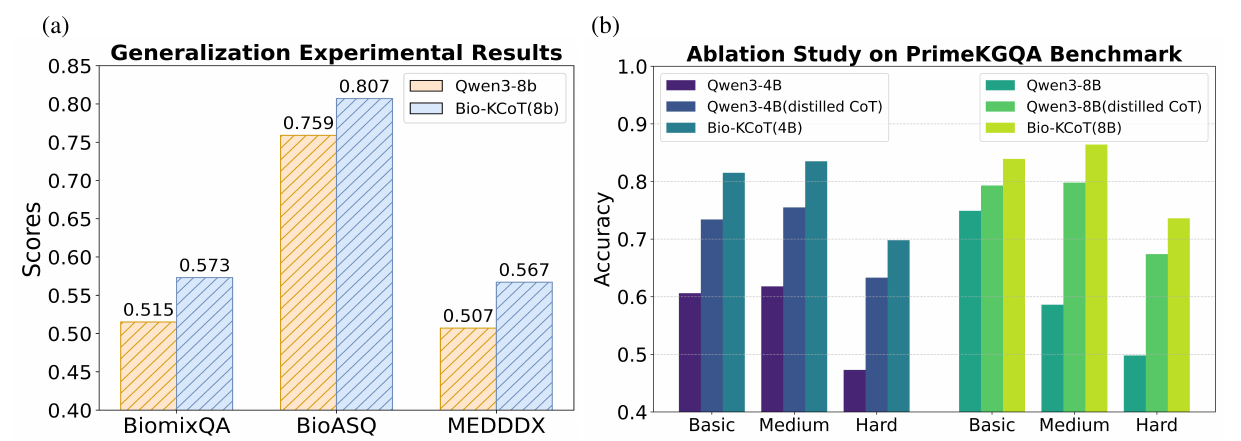}
    \caption{(a) Generalization results on additional biomedical benchmarks(BiomixQA, BioASQ, and MEDDDX). (b) Ablation study on the PrimeKGQA benchmark against the distilled CoT baseline.}
    \label{fig:generalization}
\end{figure}

{To systematically evaluate the cross-task generalization ability of our method, we further conduct experiments on three external biomolecular question-answering datasets (results shown in Figure~\ref{fig:generalization}a). We directly evaluate the model on these out-of-distribution datasets without any further training. These benchmarks span different dimensions of biomolecular challenges: BioASQ~\citep{krithara2023BioASQ} reflects real expert information needs based on MEDLINE; BiomixQA~\citep{soman2024biomedical} consists of multiple-choice questions derived from SPOKE disease–gene associations to assess complex relational reasoning; and MEDDDX~\citep{su2024knowledge} leverages STaRK-Prime~\citep{wu2024stark} to construct closely related distractors for evaluating fine-grained semantic discrimination.}

{Across these mutually distinct tasks, Bio-KCoT(8B) consistently achieves significant improvements over the base model Qwen3-8B. Notably, Bio-KCoT requires neither task-specific training nor complete external retrieval. Instead, our approach leverages knowledge graph-guided long chains of thought to effectively activate the model’s intrinsic reasoning capacity. This mechanism allows the model to derive reliable answers without depending on fully comprehensive external knowledge, while stably transferring its capabilities to new tasks and domains. These results demonstrate that long CoT not only enhances reasoning performance on in-domain tasks but also exhibits stronger cross-task generalization, laying the foundation for deployment in complex real-world scenarios.}


\subsection{Ablation Study}



{We conduct ablation studies to analyze the effectiveness of the proposed knowledge-guided long chain-of-thought generation method. Specifically, we compare the following approaches: (1) the base model Qwen3, on which all our methods are built; 
(2) Qwen3 (distilled CoT), where the supervised fine-tuning stage employs simple reasoning chains distilled from LLMs rather than knowledge-guided ones;
and (3) our proposed Bio-KCoT method. All three approaches are evaluated with models of 4B and 8B parameters.}

The experimental results are shown in Figure~\ref{fig:generalization}b. 
{Analysis of the experimental results reveals a distinct performance hierarchy that validates our proposed approach. While applying a two-stage training paradigm (SFT and RL) with distilled CoT data yields a significant performance uplift compared to the original Qwen3 base model, our model, trained with Bio-KCoT, consistently and substantially outperforms this distilled CoT counterpart across all task difficulties. Crucially, the superior performance of our model validates the efficacy of our proposed KG-guided reasoning framework, demonstrating its advanced capability in navigating complex, multi-step problems.}
See the Appendix~\ref{app:detailed_results} for detailed results.

\section{Conclusion}

We introduce Bio-KCoT, a knowledge-augmented long chain-of-thought reasoning framework for complex biomolecular problems. Our approach tackles the issues of unreliable knowledge and logical errors in large language models by using structured knowledge to guide reasoning, without depending on complete external information. Specifically, Bio-KCoT employs a knowledge graph–guided path search and pruning strategy to construct biologically meaningful and logically coherent reasoning chains. Further, we introduce PrimeKGQA, a benchmark dataset covering diverse biomolecular task types with various difficulties.
Experimental results demonstrate that Bio-KCoT improves performance on this benchmark, particularly on hard-level questions, and maintains strong generalization on out-of-distribution datasets without additional training. 

\bibliography{iclr2026_conference}
\bibliographystyle{iclr2026_conference}

\clearpage

\appendix
\section{Appendix}

\subsection{Detailed GRPO objective and implementation}
\label{app:grpo}
For a training input $x$, we sample $G$ candidate responses $\{y_i\}_{i=1}^{G}$ from the old policy $\pi_{\theta_{\text{old}}}$, and optimize the updated policy $\pi_\theta$ via the following clipped surrogate objective:
\begin{equation} \label{eq:grpo}
\begin{aligned}
\mathcal{J}(\theta) = 
\mathbb{E}_{x\sim\mathcal D,\;\{y_i\}_{i=1}^{G}\sim \pi_{\theta_{\text{old}}}(\cdot \mid x)}
\;\frac{1}{G}\sum_{i=1}^{G} \Big[
   & \min\!\big(p_i(\theta)A_i,\; 
     \operatorname{clip}(p_i(\theta),\, 1-\epsilon,\, 1+\epsilon)\,A_i\big) \\
   &\;-\; \beta \,\mathbb{D}_{\mathrm{KL}}\!\big(\pi_\theta \,\|\, \pi_{\theta_{\text{ref}}}\big) 
\Big],
\end{aligned}
\end{equation}
where $\,p_i=\dfrac{\pi_\theta(y_i\mid x)}{\pi_{\theta_{\text{old}}}(y_i\mid x)}\,$ is the probability ratio.

The normalized advantage $A_i$ is computed within the sampled group:
\begin{equation}
A_i = \frac{\,r_i - \operatorname{mean}(\{r_j\}_{j=1}^{G})\,}{\operatorname{std}(\{r_j\}_{j=1}^{G})}\,,
\end{equation}
with $r_i$ the reward of the $i$-th response. The KL penalty term $\mathbb{D}_{\mathrm{KL}}$ stabilizes training against the reference policy $\pi_{\theta_{\text{ref}}}$, where $\epsilon$ is the clipping threshold and $\beta$ the penalty weight.

Here, we provide a detailed breakdown of the components in the objective function. In this formulation, $\mathcal{D}$ is the distribution of training prompts $x$, and $G$ is the number of candidate responses sampled per prompt. The policy being optimized is $\pi_\theta$, with parameters $\theta$, while $\pi_{\theta_{\text{old}}}$ is a fixed, older version of the policy used for sampling. The reference policy, $\pi_{\theta_{\text{ref}}}$, is used to regularize $\pi_\theta$. The term $p_i(\theta)$ is the importance sampling ratio for the $i$-th response, and $A_i$ is the normalized advantage of that response, calculated using the scalar reward $r_i$. The $\operatorname{clip}(\cdot)$ function constrains the probability ratio to stabilize training. The Kullback-Leibler (KL) divergence, $\mathbb{D}_{\mathrm{KL}}(\pi_\theta \,\|\, \pi_{\theta_{\text{ref}}})$, acts as a penalty term, with its strength controlled by the coefficient $\beta$. Finally, $\epsilon$ is the clipping hyperparameter for the surrogate objective.

\subsection{Evaluation Datasets}
\label{app:datasets}
\begin{figure}[h]
    \centering
    \begin{minipage}{0.48\linewidth}
        \centering
        \includegraphics[width=\linewidth, trim=0 0 0 0, clip]{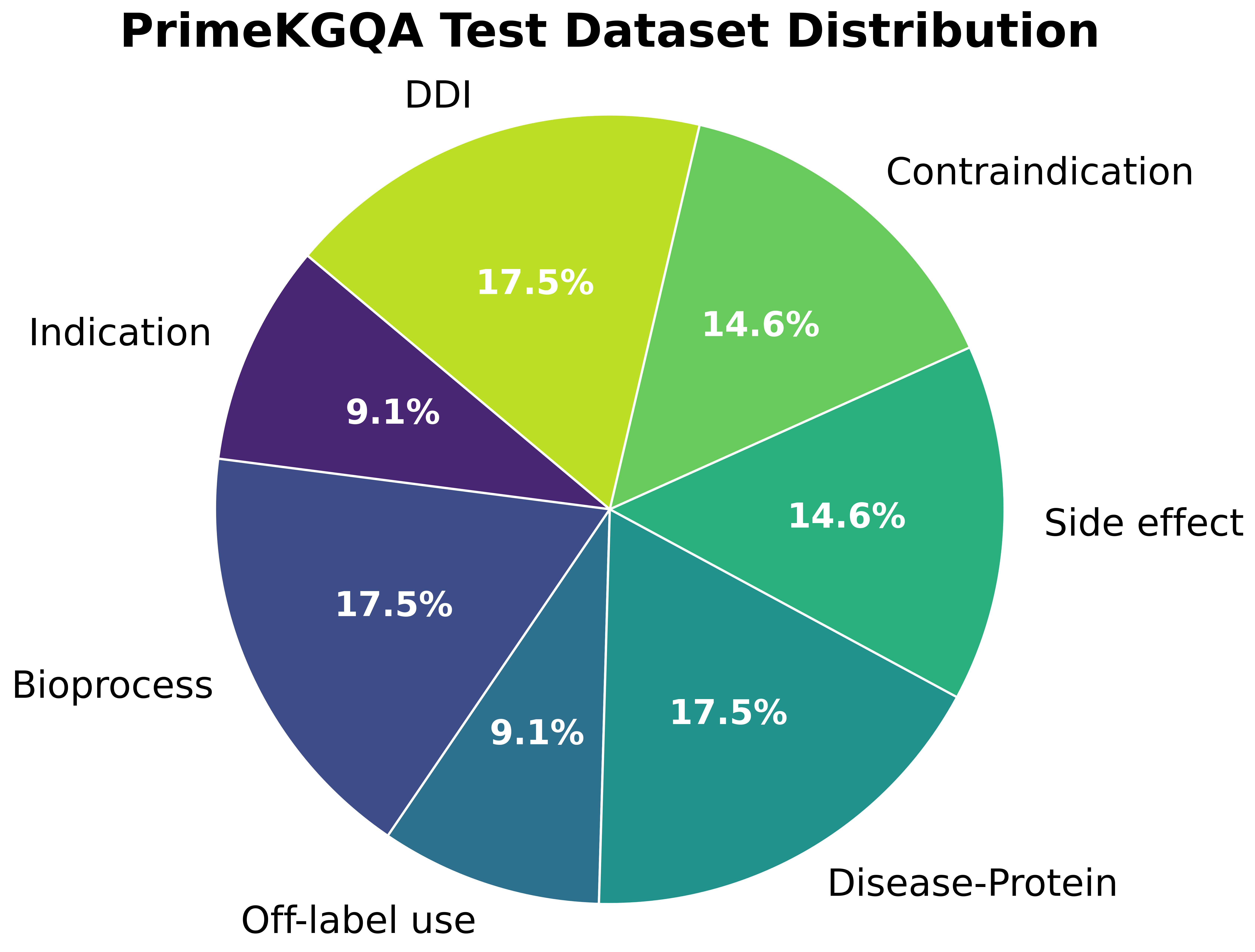}
        \caption{The distribution of the PrimeKGQA test dataset across various tasks.}
        \label{fig:data_test}
    \end{minipage}
    \hfill
    \begin{minipage}{0.48\linewidth}
        \centering
        \captionof{table}{Comprehensive statistics for our evaluation data, detailing the distribution of question types and counts across the primary test set and generalization benchmarks.}
        \label{tab:all_test_data}
        \begin{tabular}{l cc}
        \toprule
        \textbf{Dataset} & \textbf{Question Type} & \textbf{Number}\\
        \midrule
        \makecell[l]{PrimeKGQA}     & MCQ & 1710 \\
        \makecell[l]{BiomixQA} & T/F Question & 498  \\
        \makecell[l]{BioASQ}     & MCQ & 246 \\
        \makecell[l]{MEDDDX}     & MCQ & 245 \\
        \bottomrule
        \end{tabular}
    \end{minipage}
\end{figure}

The distribution of the PrimeKGQA test dataset across various tasks is presented in Figure~\ref{fig:data_test}. The number of test samples for each task is proportional to the volume of corresponding data in the source KG, thus aligning the evaluation with the inherent data distribution of the KG. Table~\ref{tab:all_test_data} presents a quantitative overview of our test data, illustrating the statistical distribution of samples between the primary test set and the generalization benchmarks. The summary specifically delineates the volume and variety of type of questions within each data set, establishing a clear and comprehensive foundation for our performance evaluation.

\subsection{Case Study and related prompts}
\label{app:prompts}

\begin{figure}[h]
    \centering    \includegraphics[width=\linewidth, trim=40 0 100 0, clip]{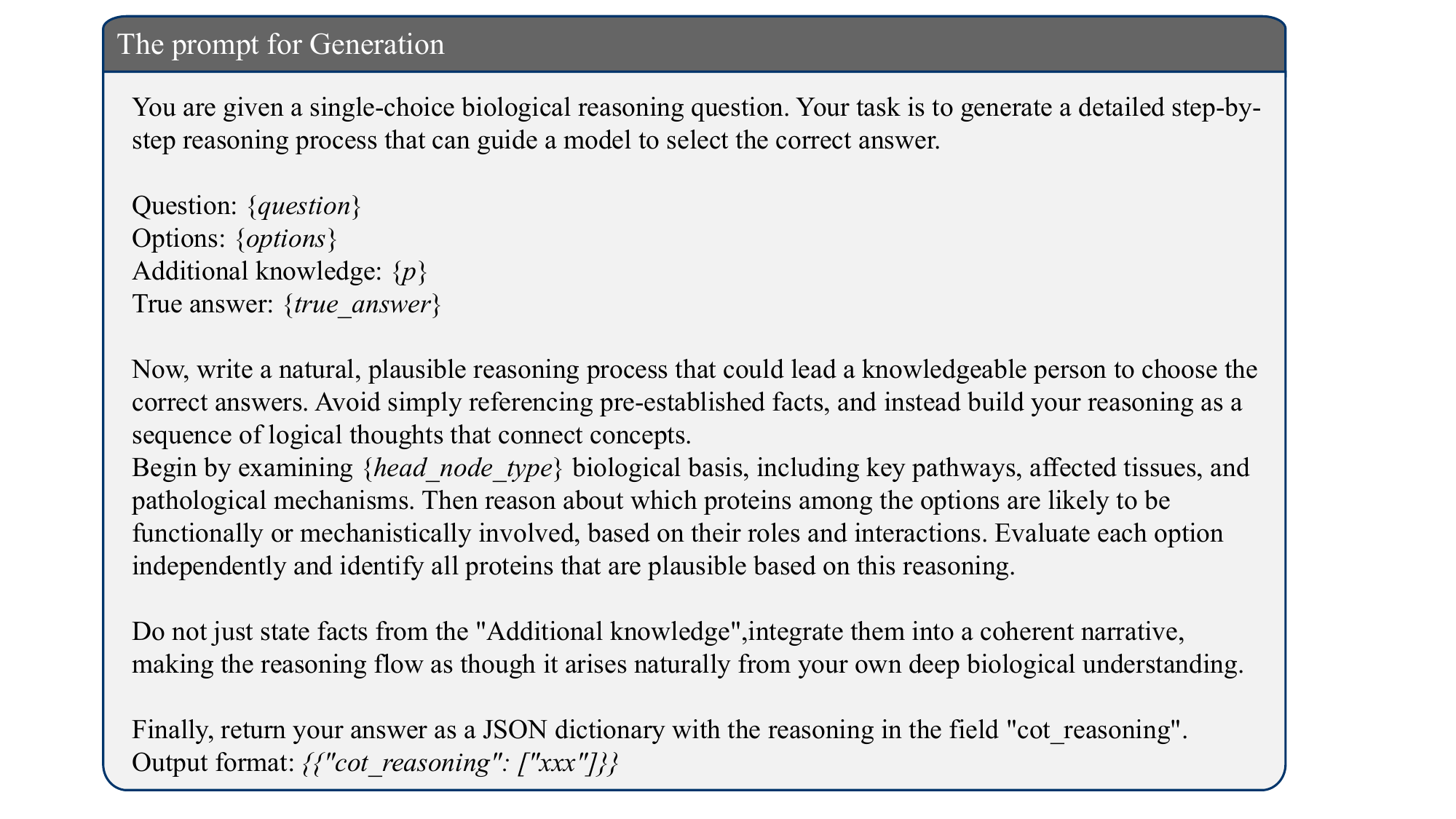}
    \caption{The prompt structure used for CoT generation. This prompt guides the model to produce a detailed, step-by-step reasoning process that leads to an initial answer based on the extracted KG path.}
    \label{fig:cot-generation}
\end{figure}

\begin{figure}[h]
    \centering    \includegraphics[width=\linewidth, trim=70 30 120 0, clip]{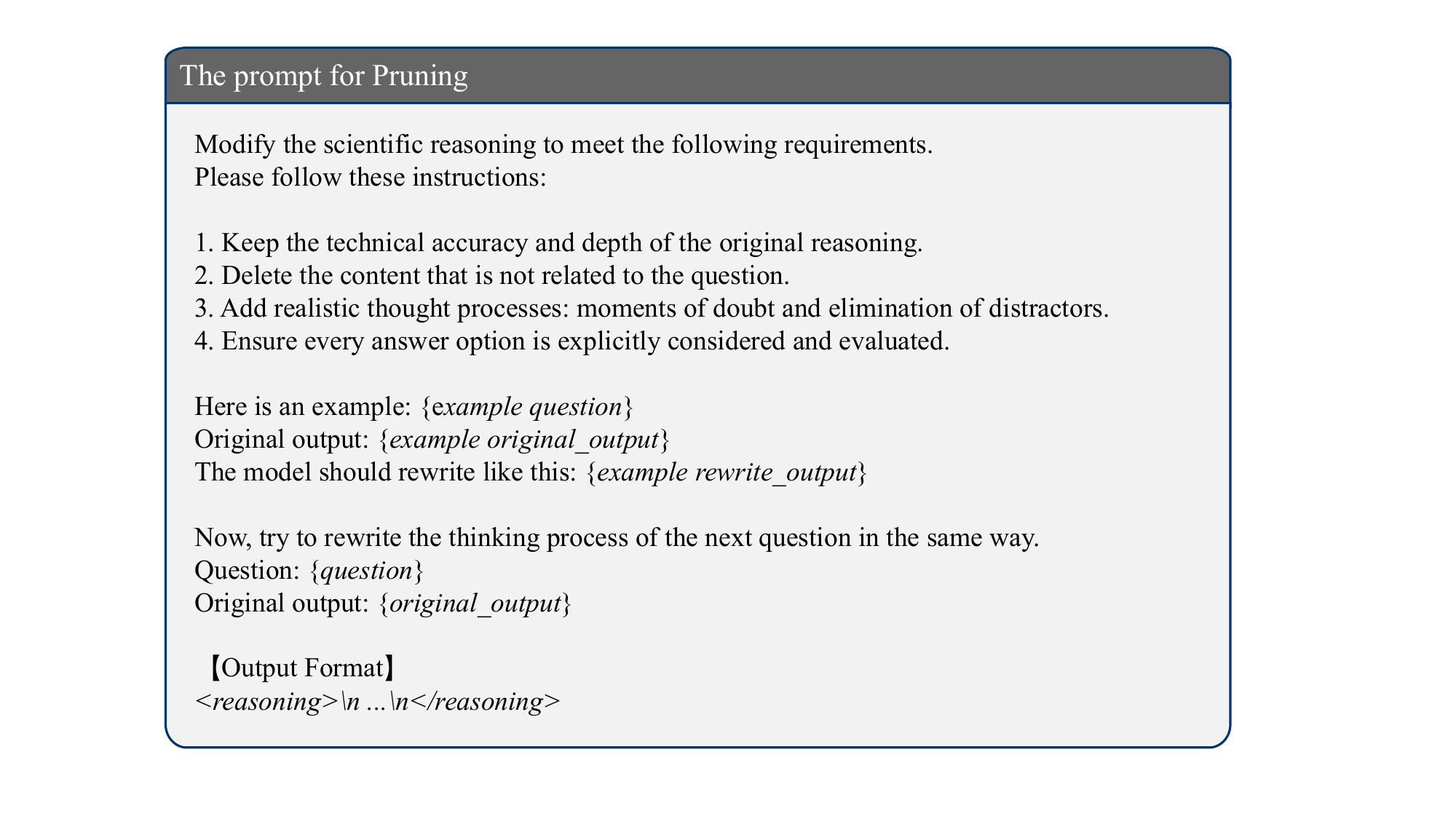}
    \caption{The prompt structure used for CoT pruning. This prompt instructs the model to refine the initial, verbose reasoning by eliminating redundancies, correcting logical failures, and distilling the core inferential steps into a final, concise answer.}
    \label{fig:cot-pruning}
\end{figure}

To provide a concrete illustration of our methodology, we present a detailed case study (Figure~\ref{fig:example}).  The process begins with the initial problem, from which we extract a relevant reasoning path within the KG. This path serves as the foundation for generating an initial, verbose response using our CoT generation prompt (Figure~\ref{fig:cot-generation}). Subsequently, this preliminary answer undergoes a refinement phase, where a specialized CoT pruning prompt is applied to revise the essential information and produce the final, concise answer (Figure~\ref{fig:cot-pruning}). Throughout these figures, we use color-coding to highlight the transformation.  Specifically, the text marked in red indicates superfluous content that deviates from a true inferential path. Such content is revised because it is either superfluous or fails to demonstrate true reasoning. This failure is evident when the model abandons the step-by-step derivation, instead opting to provide external facts directly or flagging its input with terms like "Additional knowledge". In contrast, the content marked in green represents the final, polished output, which is composed of the valid inferential chain that has been preserved and, in some cases, augmented with further elaborations to enhance logical coherence and completeness.

\begin{figure}[t]
    \centering
    \includegraphics[width=\linewidth, trim=35 55 25 35, clip]{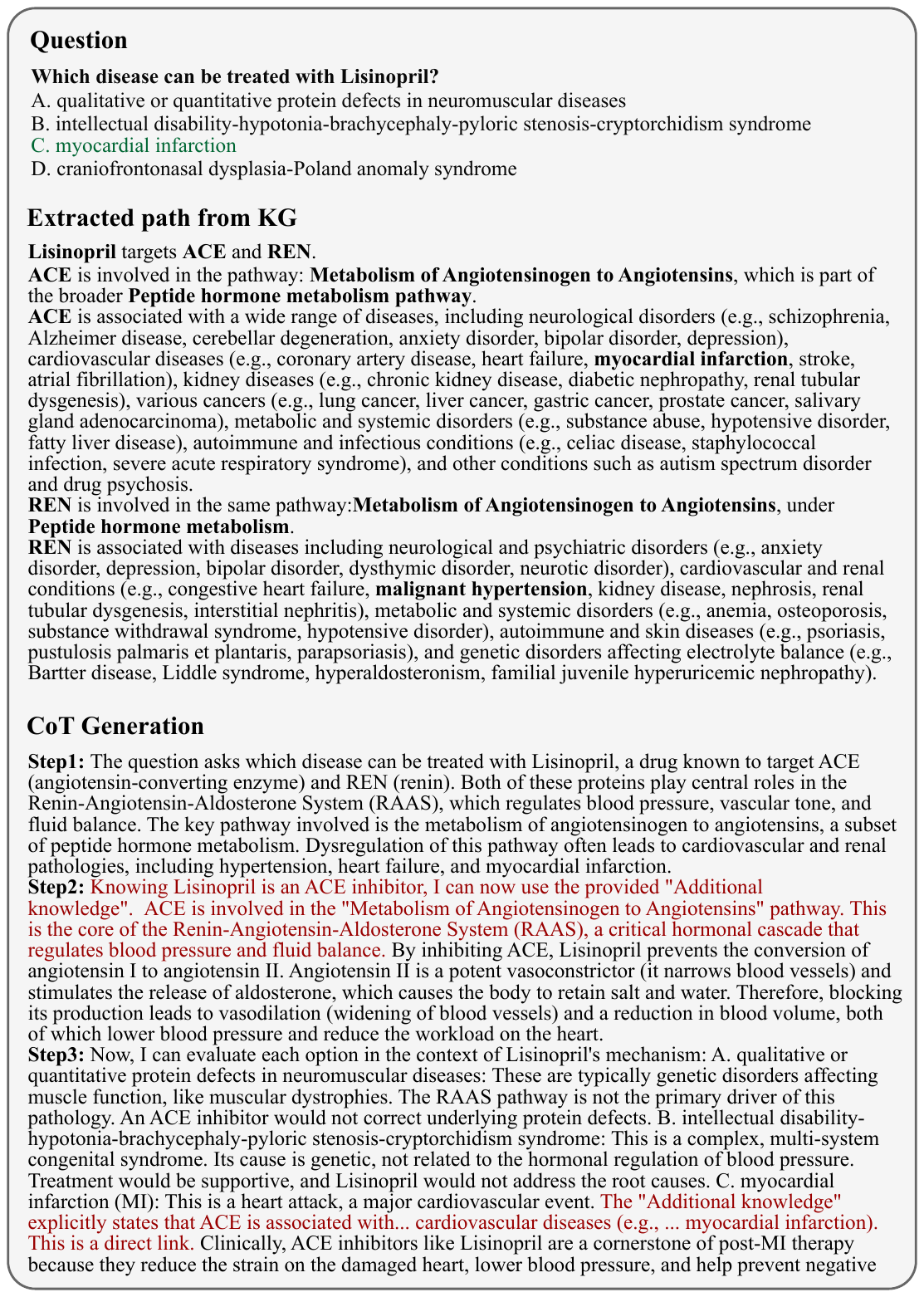}
    \caption{A detailed case study illustrating our three-stage methodology. The example demonstrates the transformation of a response, starting from the foundational KG path, to a verbose answer via CoT generation, and concluding with a concise, refined answer after CoT pruning.}
    \label{fig:example}
\end{figure}

\begin{figure}[t]\ContinuedFloat
    \centering
    \includegraphics[width=\linewidth, trim=35 35 25 30, clip]{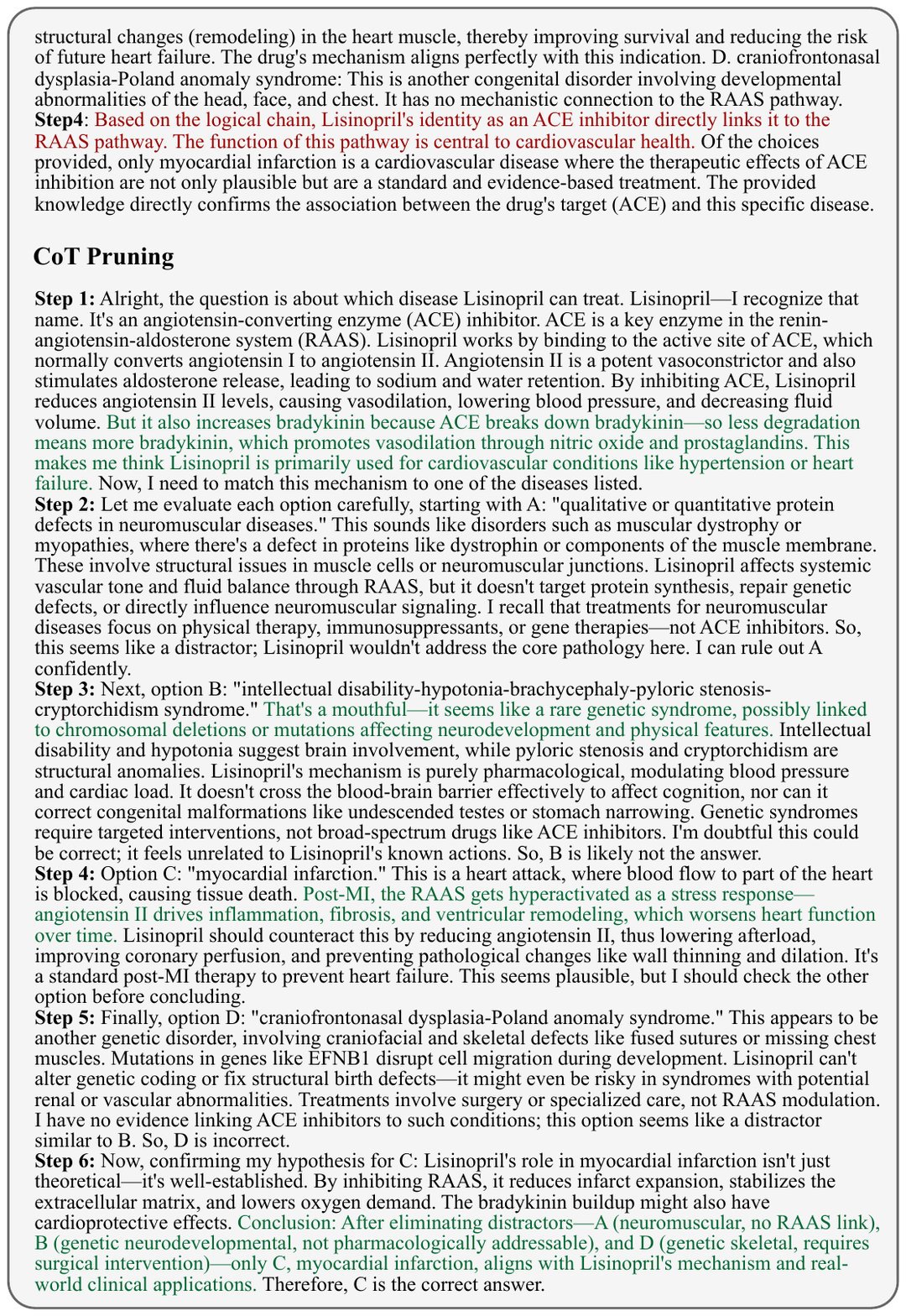}
    \caption[]{(continued)}
\end{figure}

\subsection{Detailed results}
\label{app:detailed_results}
Our experimental results comprehensively validate the effectiveness and robustness of the proposed Bio-KCoT framework. The ablation study, detailed in Table~\ref{tab:ablation-merged}, demonstrates that our method consistently and substantially outperforms both the base model and a strong distilled CoT baseline across all task categories and difficulty levels on the PrimeKGQA benchmark. This highlights the significant advantage of integrating structured knowledge into the reasoning process. Furthermore, as shown in Table~\ref{tab:model-variants-perf}, the performance gains are not confined to our primary dataset; the model maintains its superiority on several external biomedical benchmarks. This confirms that the enhanced reasoning capabilities generalize effectively to unseen data and diverse problem types, establishing Bio-KCoT as a powerful and broadly applicable framework for complex biomedical question answering.

\begin{table}[h]
\centering
\caption{Ablation study on PrimeKGQA. Experiments are performed across all tasks with the 4B and 8B model variants, in comparison with the distilled CoT baseline.}
\label{tab:ablation-merged}
\renewcommand{\arraystretch}{1.2} 
\resizebox{\textwidth}{!}{
\begin{tabular}{l cc c ccc c cc c c}
\toprule
\multirow{2}{*}[-1.5ex]{\textbf{Model}} & \multicolumn{3}{c}{\textbf{Basic}} & \multicolumn{4}{c}{\textbf{Medium}} & \multicolumn{3}{c}{\textbf{Hard}} & \multirow{2}{*}[-1.5ex]{\textbf{All Avg.}} \\
\cmidrule(lr){2-4} \cmidrule(lr){5-8} \cmidrule(lr){9-11}
& \thead{Indi- \\ cation} & \thead{Bio- \\ process} & Avg. & \thead{Off-label \\ use} & \thead{Disease- \\ Protein} & \thead{Side \\ effect} & Avg. & \thead{Contra- \\ indication} & DDI & Avg. & \\
\midrule
\makecell[l]{Qwen3-4B} & 0.768 & 0.523  & 0.606 & 0.710 & 0.573  & 0.616 & 0.618 & 0.438 & 0.503 & 0.473 & 0.568 \\
\makecell[l]{Qwen3-4B \\ (distilled CoT)} & 0.819  & 0.690 & 0.734 & 0.800 & 0.687 & 0.808 & 0.755 & 0.720 & 0.560 & 0.633 & 0.710 \\
\makecell[l]{\textbf{Bio-KCoT(4B)}} & \underline{0.897}  & \underline{0.773}  & \underline{0.815} & \underline{0.890} & 0.743    & \textbf{0.912} & \underline{0.835} & \underline{0.804} & \underline{0.610} & \underline{0.698} & \underline{0.786}  \\
\makecell[l]{Qwen3-8B} & 0.845 & 0.653 & 0.749 & 0.671 & 0.600 & 0.516 & 0.586 & 0.452 & 0.537 & 0.498 & 0.601 \\
\makecell[l]{Qwen3-8B \\ (distilled CoT)} & 0.869 & 0.753 & 0.793 & 0.851 & \underline{0.747} & 0.828 & 0.798 & 0.768 & 0.597 & 0.674 & 0.773 \\
\makecell[l]{\textbf{Bio-KCoT(8B)}} & \textbf{0.901}  & \textbf{0.807} & \textbf{0.839} & \textbf{0.930} & \textbf{0.793}    & \underline{0.908} & \textbf{0.864} & \textbf{0.848} & \textbf{0.643} & \textbf{0.736} & \textbf{0.816} \\
\bottomrule
\end{tabular}
}
\end{table}

\begin{table}[h]
\caption{Generalization study on three biomedical benchmarks (BiomixQA, BioASQ, and MEDDDX). Performance is evaluated using the 4B and 8B model variants.}
\label{tab:model-variants-perf}
\centering
\begin{tabular}{l ccc}
\toprule
\textbf{Model} & \textbf{BiomixQA} & \textbf{BioASQ} & \textbf{MEDDDX} \\
\midrule
\makecell[l]{Qwen3-4b}     & 0.472 & 0.659 & 0.426 \\
\makecell[l]{\textbf{Bio-KCoT(4b)}} & 0.512(\textbf{+4.0}) & 0.727(\textbf{+6.8})
& 0.473(\textbf{+4.7}) \\
\makecell[l]{Qwen3-8b}     & 0.515 & 0.759 & 0.507 \\
\makecell[l]{\textbf{Bio-KCoT(8b)}} & 0.573(\textbf{+5.8}) 
& 0.807(\textbf{+4.8}) 
& 0.567(\textbf{+6.0}) \\
\bottomrule
\end{tabular}
\end{table}

\subsection{Error Analysis}
\label{app:error}
\begin{figure}[h]
    \centering    \includegraphics[width=\linewidth, trim=35 120 25 35, clip]{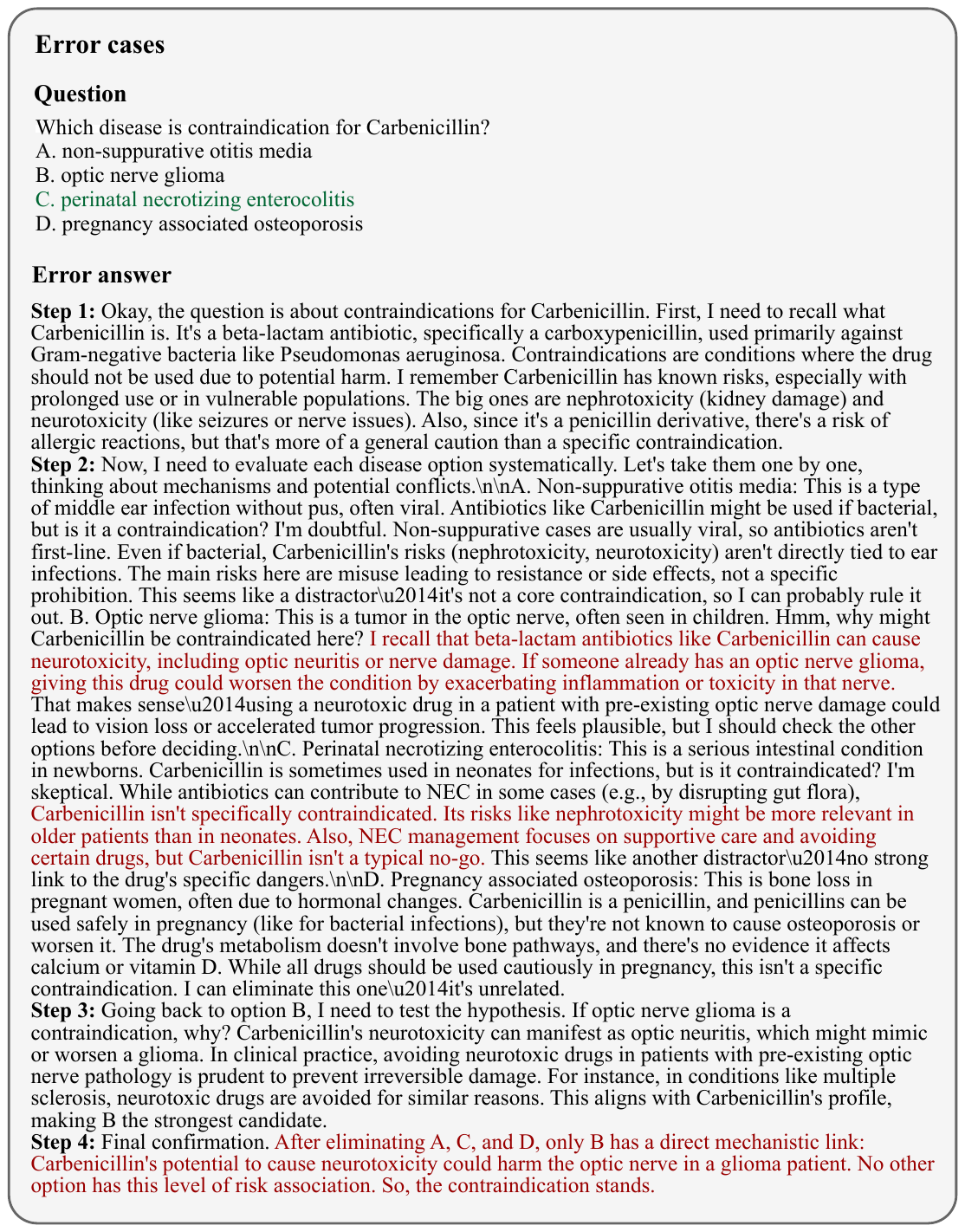}
    \caption{An illustrative example of the error analysis process for a failure case. The error is traced from the final answer back through the reasoning chain, revealing that the failure originates from a knowledge deficit rather than a flaw in logical deduction.}
    \label{fig:error}
\end{figure}

While the Bio-KCoT framework demonstrates substantial advancements in reasoning performance, a systematic error analysis is essential for diagnosing its inherent limitations and guiding future iterations. This section presents a detailed error analysis of failures arising during the generation and execution of reasoning chains. Errors are traced from the final answers (highlighted in red) back through the CoT generation and execution steps (Figure~\ref{fig:error}). This process helps to illuminate the model's current limitations, and informing pathways for future refinement. The analysis indicates that most errors do not arise from flawed logical deduction but are instead attributable to the limited knowledge capacity of smaller models, which may lead to errors when recalling specific facts or specialized knowledge points. Consequently, even when the reasoning process is internally consistent, the final outputs can contain inaccuracies or misconceptions, highlighting the importance of combining reasoning analysis with knowledge verification in future iterations.

\section{Ethics Statement}
This work complies with the ICLR Code of Ethics. Our study did not involve human participants or animal testing. All datasets, including PrimeKGQA, were obtained and used in accordance with relevant guidelines, ensuring respect for data usage policies and privacy considerations. We carefully mitigated potential biases and avoided discriminatory effects during the research process. No personal or identifiable information was utilized, and no experiments were carried out that might raise concerns regarding security or confidentiality. Throughout the study, we have upheld principles of transparency, responsibility, and research integrity.

\section{Reproducibility Statement}
To promote transparency and ensure reproducibility, we have provided all relevant code in the Supplementary Material. The manuscript offers a comprehensive description of the experimental setup, including the training pipeline, model architectures, hyperparameter configurations. These materials are intended to enable independent verification of our findings and to support subsequent research that builds upon our methodology.

\section{The use of Large Language Models}
Large Language Models were employed as versatile assistive tools in this research. They were primarily used to support and refine the preparation of this manuscript, including correcting grammar, enhancing clarity, and restructuring sentences to improve overall readability. LLMs also assisted in generating foundational scripts for data preprocessing tasks such as cleaning, formatting, and visualization, which were subsequently reviewed and adjusted by the authors to ensure correctness and consistency with research objectives.

In addition, LLM APIs were invoked during the CoT generation and pruning stages, where they contributed to producing candidate reasoning traces and improving their quality through refinement. Across all these applications, the outputs provided by LLMs were critically assessed and edited by the authors. At no stage did the models substitute for human intellectual contributions; rather, they served to accelerate and augment different phases of the research workflow.

\end{document}